# Leveraging Big Data Frameworks for Spam Detection in Amazon Reviews

Mst Eshita Khatun*, Halima Akter*, Tasnimul Rehan*, Toufiq Ahmed†
*Department of CSE, Daffodil International University, Dhaka, Bangladesh
†Department of CSE, American International University Bangladesh, Dhaka, Bangladesh
Emails: eshita.cse@diu.edu.bd, halima15-13219@diu.edu.bd, tasnimul15-13216@diu.edu.bd, 22-47299-1@student.aiub.edu

***Abstract*—In this digital era, online shopping is common practice in our daily lives. Product reviews significantly influence consumer buying behavior and help establish buyer trust. However, the prevalence of fraudulent reviews undermines this trust by potentially misleading consumers and damaging the reputations of the sellers. This research addresses this pressing issue by employing advanced big data analytics and machine learning approaches on a substantial dataset of Amazon product reviews. The primary objective is to detect and classify spam reviews accurately so that it enhances the authenticity of the review. Using a scalable big data framework, we efficiently process and analyze a large scale of review data, extracting key features indicative of fraudulent behavior. Our study illustrates the utility of various machine learning classifiers in detecting spam reviews, with Logistic Regression achieving an accuracy of 90.35%, thus contributing to a more trustworthy and transparent online shopping environment.**

## I. Introduction

The exponential growth of the Internet has profoundly transformed daily human activities and has offered convenience across various domains of life, including communication and e-commerce. In the context of e-commerce, customer reliance on online product reviews has become a pivotal factor in shaping purchasing decisions. Writing reviews is a common feature on most e-commerce platforms that allows users to share valuable feedback based on their experiences. These reviews help assess vendor credibility and product authenticity without direct physical evaluation. However, this dependence also creates opportunities for fraudulent manipulation [1]. The continuous growth and complexity of online marketplaces have made it challenging for consumers to distinguish authentic reviews, thereby increasing their vulnerability to misleading and deceptive content. Therefore, only an authentic review can gain customer trust and this will reduce the rate of customer fraud. During the pandemic, around 200 million spam reviews were detected on Amazon, and approximately ten thousand Facebook groups were found promoting misleading reviews [2]. With 5.8 million Amazon reviews from 2.14 million reviewers, identifying spam reviews through conventional market research becomes challenging, as unethical traders frequently buy fake positive reviews or employ automated bots to artificially boost product visibility [3]. Such misleading spam reviews interrupt the faith between the buyer and the vendor. To mitigate this issue, our goal is to classify reviews and identify spam reviews to ensure safer purchasing. Furthermore, establishing robust detection methods can significantly contribute to maintaining the integrity and transparency of e-commerce platforms.

This study aims primarily to analyze a large data set of 3.5 million records within a big data framework. Leveraging big data technologies is essential in this context, as traditional data processing tools cannot adequately manage the massive data volumes of e-commerce platforms. The distributed processing capabilities of big data frameworks, such as PySpark, enable efficient computation, scalability, and real-time insights across massive datasets [4]. The study aims to identify spam reviewers through time series analysis and behavioral segmentation, as well as detect spam reviews using machine learning (ML) algorithms. Furthermore, pre-processing steps such as tokenization, stop word removal, and TF-IDF transformation are incorporated to refine input features and support more accurate model predictions.

The subsequent sections present the related work, followed by a comprehensive overview of data pre-processing, feature extraction, and algorithm implementation in Section III. In Section IV discussion of result, and finally conclusion.

## II. Related Works

In the related works section, we reviewed prior studies to understand the significance of spam detection in online platforms. Both studies [5] and [6] explored spam detection using ML algorithms and NLP techniques. Fayaz et al. [7] proposed an ensemble approach combining MLP, KNN, and Random Forest, achieving a classification accuracy of 89.26% on the Yelp dataset. Shahariar et al. [8] presented a spam review detection technique leveraging deep learning with labeled and unlabeled datasets, and LSTM outperforms traditional machine learning classifiers.In a similar objective, Mani et al. [9] developed a fake review detection system using n-gram features and combined learning algorithms that shows significant efficiency. Dhingra et al. [10] utilized fuzzy model-based solutions, such as fuzzy supervised learning and the fuzzy ranking evolution algorithm for large-scale dataset analysis.

In a similar fashion, to identify fake restaurant reviews, Lee et al. [11] present 16 determinants derived from over 43,000 online reviews, with SVM suggested as their classification algorithm. Choi et al. [12] also propose a methodology to identify fake product reviews and assess their usefulness on e-commerce platforms. Their approach employs both unsupervised methods (K-means, hierarchical clustering) and supervised learning techniques (SVC and LGBM), analyzing linguistic and behavioral features of reviews. The highest classification accuracy in their study was achieved by SVC at 85.1%. In contrast, our proposed methodology and exploratory analysis of reviews offer a more comprehensive approach by incorporating large-scale data processing, reviewer behavior segmentation, and temporal analysis to enhance the accuracy and reliability of spam detection.

## III. Methodology

This section details the methodological framework adopted for the systematic identification of spam reviews within a large-scale review corpus, as illustrated in Figure 1. In our methodology, there are five main main key steps such as a) data pre-processing to improve the data quality, b) feature selection, c) exploratory data analysis to get the insight of the data properties, d) applying machine learning algorithms, and finally, e) the evaluation of detection model performance.

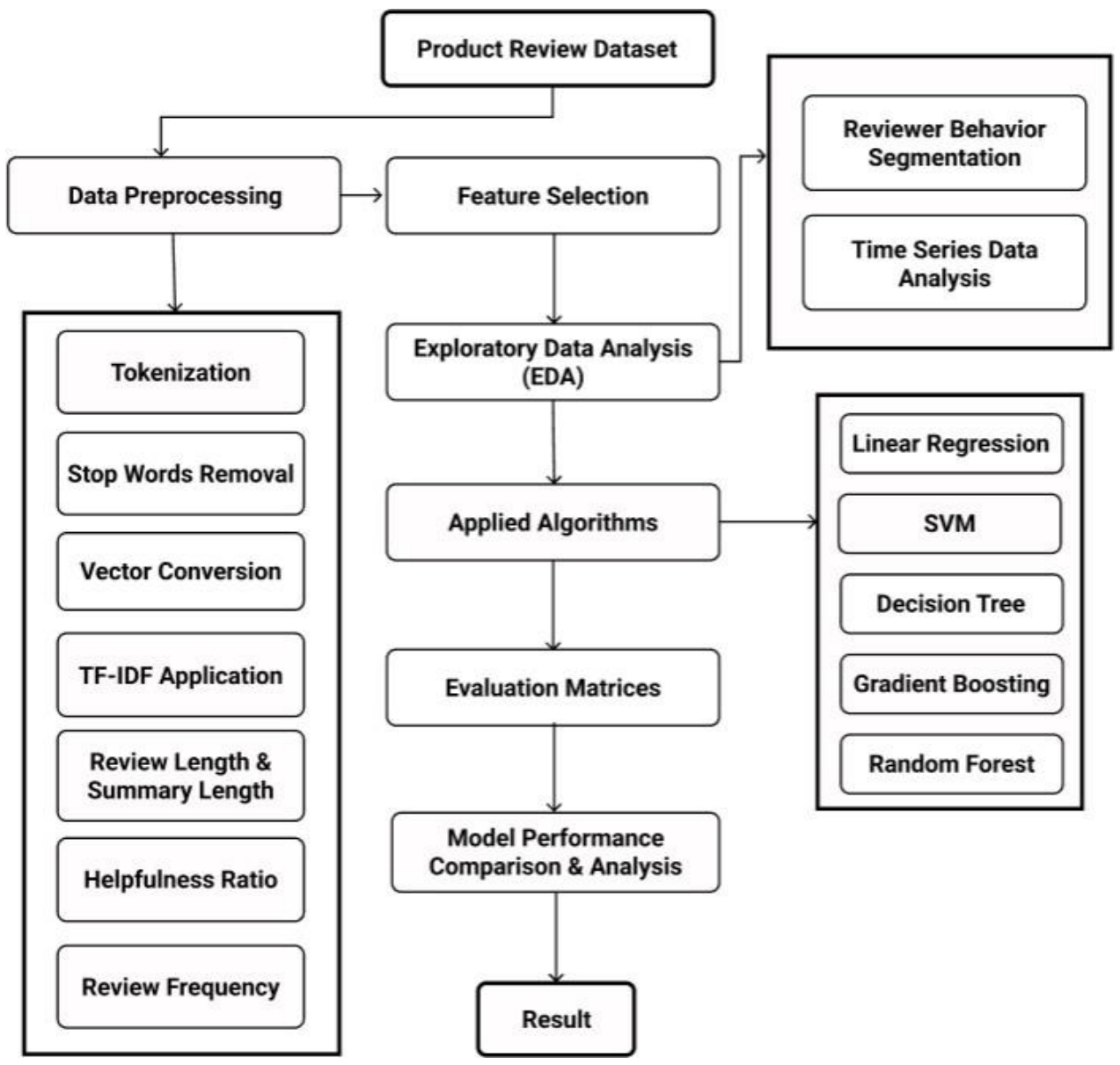

Fig. 1: Proposed Methodology for spam detection

### A. Dataset Description

In this research, we have chosen a large corpus of amazon reviews labeled dataset comprising 26.7 million Amazon reviews provided by approximately 15.4 million users in different categories of product reviews such as cellphones and accessories, clothing, shoes and jewelry, electronics, home and kitchen, sports and outdoor, and toys and games [13]. This study uses dataset of cellphones and accessories, the raw size of the dataset is 5.36 GB, and the compressed size is 1.67 GB. The dataset comprises Amazon product reviews containing unique identifiers for reviewers and products, reviewer details, review text, ratings, helpfulness votes, timestamps, and class labels indicating spam ("0") or non-spam ("1") reviews.

### B. Data pre-processing

To prepare the dataset effectively, we leveraged PySpark, a Python API for Apache Spark, which is optimized for processing large-scale datasets. PySpark provides support for distributed data processing and enables parallel computation, significantly reducing processing time for large-scale datasets while enabling efficient and scalable execution of complex data workflows. This study data pre-processing steps include:

*1) Data Cleaning:* Null values, irrelevant columns, and duplicate entries were removed to ensure data quality and consistency.

*2) Tokenization:* In text analysis, tokenization is the process of segmenting a stream of textual data into smaller units or tokens, or words, facilitating meaningful linguistic analysis. Tokenization example is presented in table I.

TABLE I: Example of Tokenization

| Step | Text Example |
|---|---|
| Original Review | "This phone case is great." |
| After Tokenization | ["This", "phone", "case", "is", "great"] |

*3) Stop word removal:* In this phase of text preprocessing, the elimination of frequently occurring yet semantically insignificant words from tokenized text refines the dataset and improves subsequent analytical precision. An example is presented in Table II.

TABLE II: Example of Stop Word Removal

| Step | Text Example |
|---|---|
| Original Tokens | ["This", "phone", "case", "is", "great"] |
| After Stop Word Removal | ["phone", "case", "great"] |

*4) Term Frequency-Inverse Document Frequency (TF-IDF):* In this phase, we utilized TF-IDF weighting to quantify the significance of terms within individual reviews in relation to their distribution across the entire corpus, thereby enhancing the relevance of extracted features for downstream classification tasks.

### C. Feature Selection

Additional numeric features were derived from raw data, such as review length, summary length, helpfulness ratio (indicating the proportion of helpful votes), and reviewer frequency metrics, thus enriching the dataset for deeper analysis. Finally, Chi-square $\chi^2$ feature selection was applied to identify the most relevant attributes to improve the efficiency and utility of the classifiers.

A correlation is calculated between the parameters that represents in Figure 2, the strong positive correlation (0.91) between the feature variable class and rating. unixReviewTime has weak positive correlations with both class and rating, but shows a moderate negative correlation with subsidiary and reviewLength.

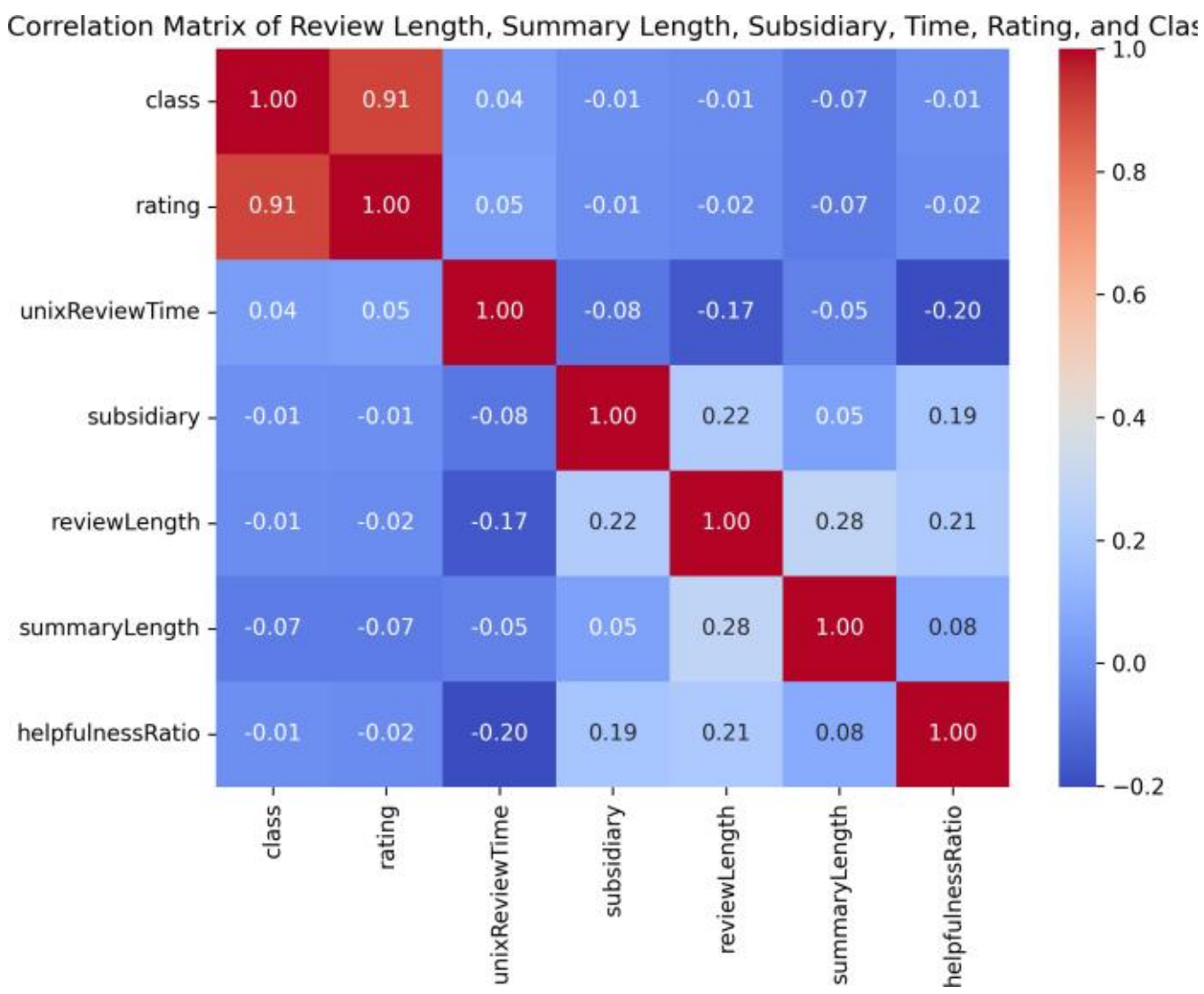

Fig. 2: Correlation between parameters

### D. Exploratory Data Analysis

This study emphasizes not only the analysis of reviews themselves but also focuses significantly on the behavioral patterns of reviewers. To systematically understand reviewer behaviors, reviewers have been categorized into three distinct groups based on their review activity frequency, as illustrated in Figure 3. These categories are 1) Frequent Reviewers, 2) Rare Reviewers, and 3) Occasional Reviewers. Among these, Rare Reviewers constitute the largest proportion, accounting for approximately 74.35% of the total.

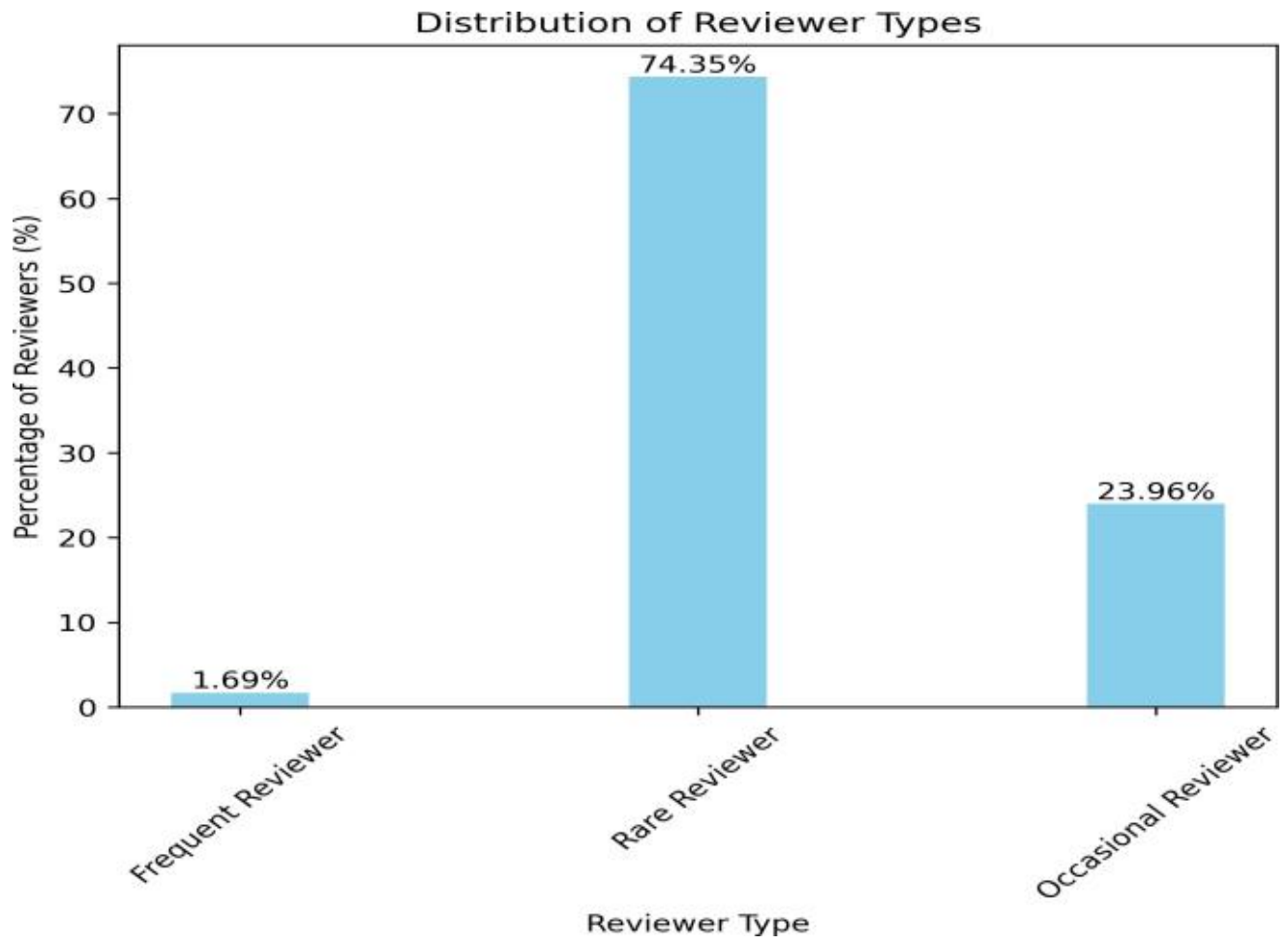

Fig. 3: Category of Reviewers

Considering reviewer behavior further, time-series analysis provides essential insights into temporal patterns, significantly contributing to the overall classification accuracy of spam reviews. Specifically, trends in average ratings provided by reviewers on a monthly and yearly basis, alongside average review length, have been visualized through Figures 4 and 5, respectively.

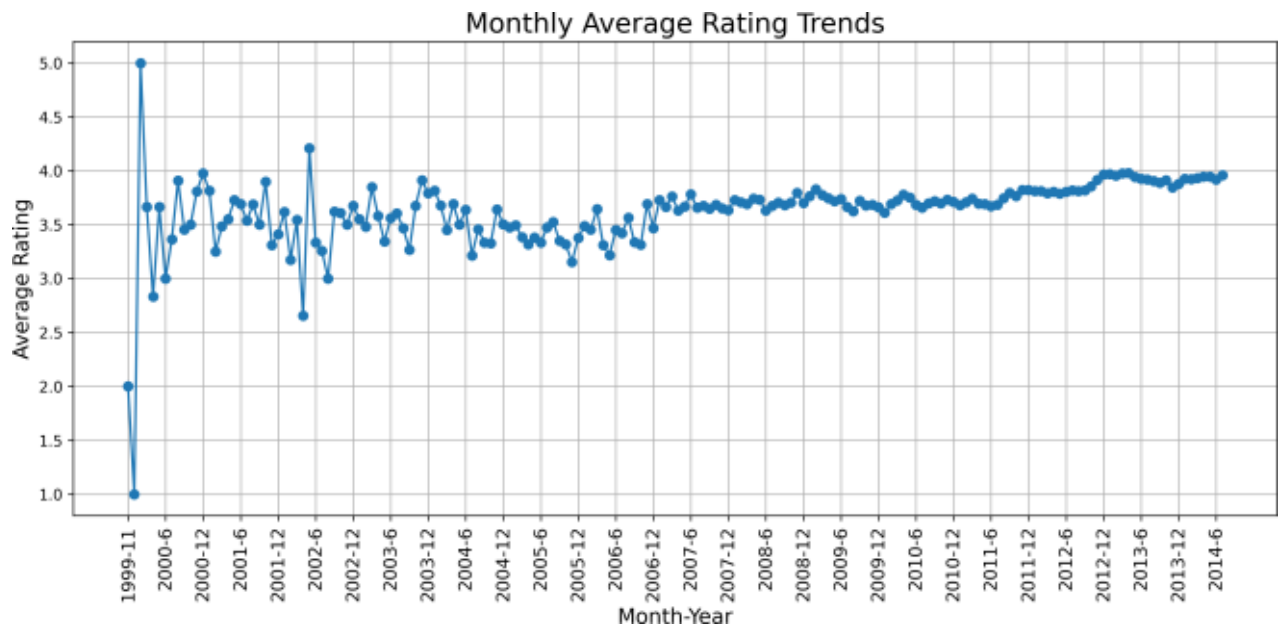

Fig. 4: Monthly Average Rating Trends

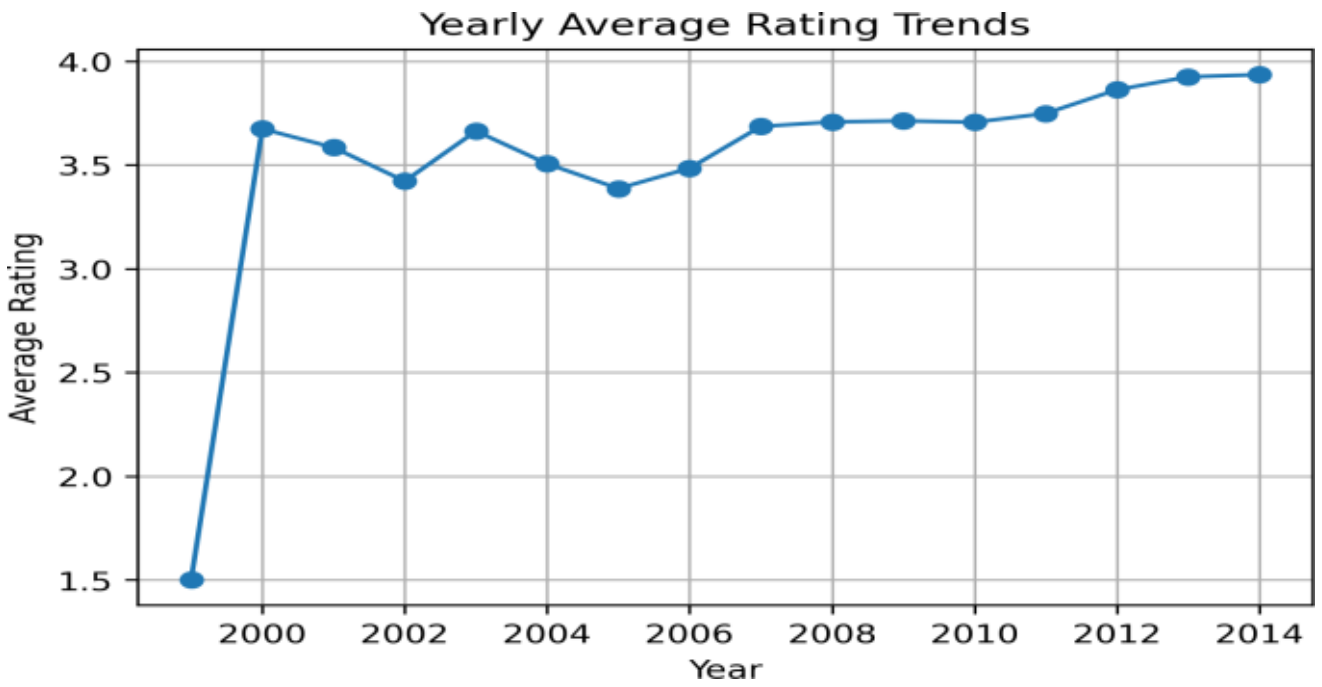

Fig. 5: Yearly Average Rating Trends

It is particularly noteworthy in Figure 5 that the average rating exhibited considerable growth over the analyzed period. For instance, the average rating, which was approximately 1.5 out of 5 in the year 2000, steadily increased, reaching close to 4.0 between 2012 and 2014. A similar upward trend can be observed in monthly average ratings.

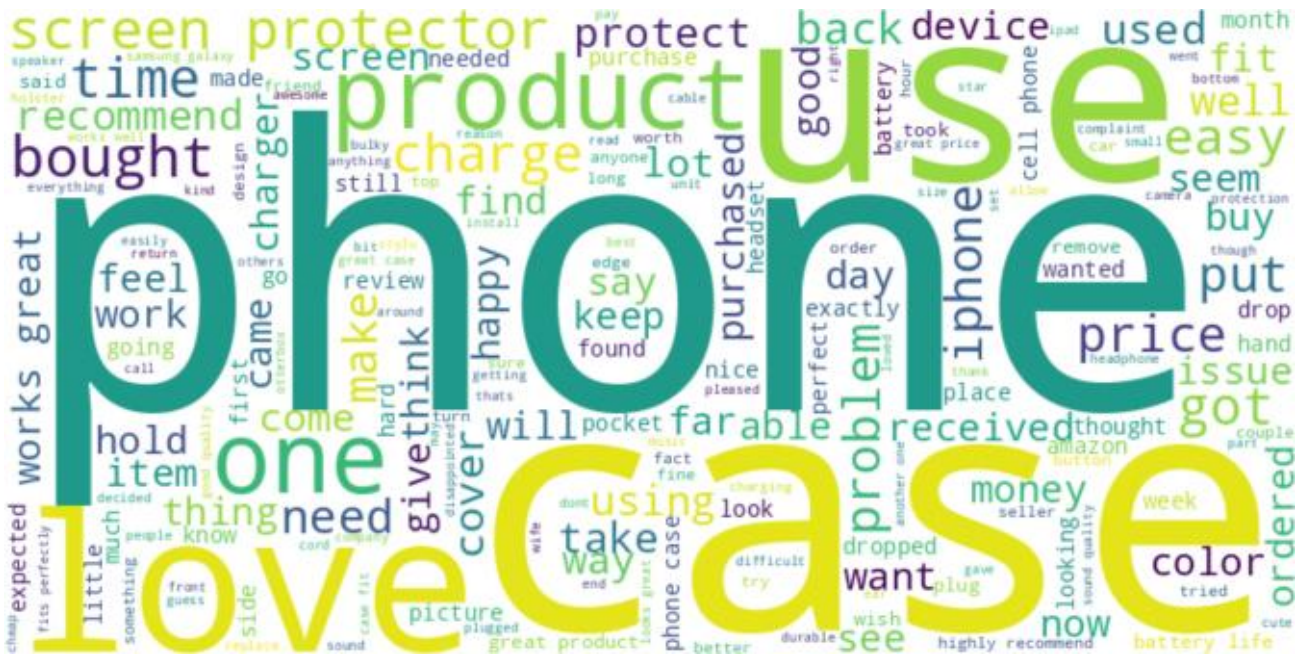

Fig. 6: Frequently Occurring Words in Amazon Reviews

Additionally, textual analysis through word cloud visualization, as depicted in Figure 6, highlights the most frequently occurring words within the review dataset. This visualization aids in identifying common themes and sentiments expressed by reviewers, thereby providing deeper qualitative insights into reviewer behavior and product perceptions.

## IV. Result and Discussion

In this research, we have employed five machine learning classification approaches, namely Logistic Regression (LR), Support Vector Machine (SVM), Random Forest (RF), Gradient Boosting (GB), and Decision Tree (DT), and spam review detection performance was assessed using a detailed evaluation based on metrics including accuracy, precision, recall, and F1-score. Collectively, these matrices provide robust insights into how well the classifiers perform in recognizing spam within extensive review datasets. Each metric depends on categorizing the review predictions into four possible outcomes. Reviews correctly identified as spam are termed true positives (TP), while reviews mistakenly marked as spam are false positives (FP). Conversely, correctly identified genuine reviews are true negatives (TN), and spam reviews incorrectly flagged as genuine are false negatives (FN).

The equation is as follows:

$$\text{Accuracy} = (TP + TN)/(TP + TN + FP + FN) \quad (1)$$

$$\text{Precision} = TP/(TP + FP) \quad (2)$$

$$\text{Recall} = TP/(TP + FN) \quad (3)$$

$$\text{F1-score} = 2 \times (\text{Precision} \times \text{Recall})/(\text{Precision} + \text{Recall}) \quad (4)$$

In Figure 7 and Table III, the LR classifier obtained the highest accuracy rate of 90.35%, demonstrating overall effectiveness in classifying Amazon reviews into spam or non-spam categories. The precision and recall scores of 90.35% and 88.09%, respectively, highlight the robust ability of the classifier to minimize false positives while effectively identifying actual spam reviews.

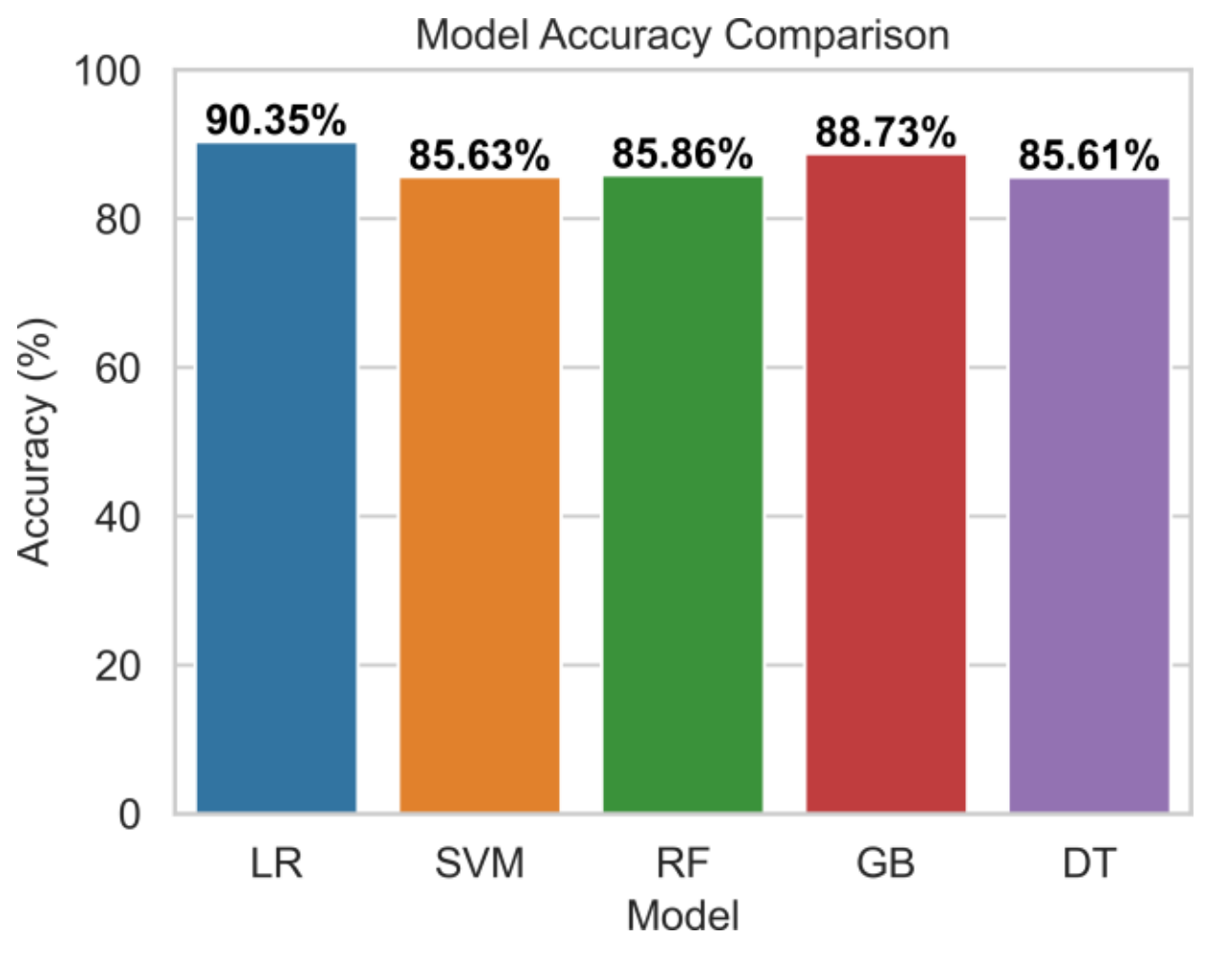

Fig. 7: Accuracy comparison of different algorithms

The SVM model reported an accuracy of 85.63%. However, it notably achieved a high recall rate of 90.35%, indicating strong proficiency in identifying spam reviews, though slightly lower precision at 88.09% suggests a higher occurrence of false positives compared to Logistic Regression. The RF classifier provided an accuracy of 85.86%, alongside the highest precision rate among all models at 92.79%. This high precision score underscores Random Forest's capability to minimize false positives significantly, although a comparatively lower recall rate of 85.86% reflects missed spam cases.

TABLE III: Performance Metrics of spam classifications models

| Model | Precision (%) | Recall (%) | F1 Score (%) |
|---|---|---|---|
| LR | 90.35 | 88.09 | 89.21 |
| SVM | 88.09 | 90.35 | 89.21 |
| RF | 92.79 | 85.86 | 89.19 |
| GB | 87.54 | 88.73 | 88.13 |
| DT | 60.57 | 85.61 | 70.95 |

Gradient Boosting showed commendable accuracy performance of 88.73%, alongside a balanced precision and recall score of approximately 87.54% and 88.73%, respectively. Lastly, DT model yielded an accuracy rate of 85.61%. Despite the comparable accuracy, its significantly lower precision score of 60.57% indicates a higher rate of false positives, considerably reducing its reliability for precise spam classification. Its recall rate of 85.61%, further highlights the trade-off between precision and recall in this specific case. In Table IV presents a comparative analysis of previous studies on spam detection, including the datasets used, the number of reviews, and the best-performing classification models along with their respective evaluation metrics.

TABLE IV: Comparison of different studies

| Reference | Review Dataset | Number of Reviews | Best Performing Model |
|---|---|---|---|
| Saumya et al. [14] | Amazon Product Review | 39,382 | Random Forest with F1-score 91% |
| Barbado et al. [15] | Online E-commerce Review | 18,192 | AdaBoost Classifier with F1-score 82% |
| Our Approach | Amazon Product Review | Over 3.5 million | Logistic Regression with 90.35% accuracy |

## V. Conclusion

In conclusion, this study leverages PySpark for scalable data pre-processing and feature engineering over 3.5 million reviews of Amazon to identify the fake product reviews. Among five machine learning classification algorithms, Logistic regression achieves the highest accuracy of 90.35%. Additionally, time series data analysis reveals seasonal trends and shifts in review behavior, offering deeper insights into reviewer dynamics. The segmentation of reviewers based on frequency further enhances the understanding of potential spam patterns in large-scale review systems. This research future work may focus on integrating deep learning models to capture contextual semantics in reviews more effectively. Additionally, incorporating multilingual datasets could broaden the applicability of the spam detection framework across diverse e-commerce platforms. Real-time deployment and adaptive learning mechanisms may also be explored to enhance responsiveness and accuracy over time.

## References

[1] S. He, B. Hollenbeck, and D. Proserpio, “The market for fake reviews,” *Marketing Science*, vol. 41, no. 5, pp. 896–921, 2022.

[2] M. Tabany and M. Gueffal, “Sentiment analysis and fake amazon reviews classification using svm supervised machine learning model,” *Journal of Advances in Information Technology*, vol. 15, no. 1, pp. 49–58, 2024.

[3] B. K. Shah, A. K. Jaiswal, A. Shroff, A. K. Dixit, O. N. Kushwaha, and N. K. Shah, “Sentiments detection for amazon product review,” in *2021 International conference on computer communication and informatics (ICCCI)*. IEEE, 2021, pp. 1–6.

[4] Y. K. Gupta and S. Kumari, “A study of big data analytics using apache spark with python and scala,” in *2020 3rd International Conference on Intelligent Sustainable Systems (ICISS)*. IEEE, 2020, pp. 471–478.

[5] H. Le and B. Kim, “Detection of fake reviews on social media using machine learning algorithms,” *Issues in Information Systems*, vol. 21, no. 1, pp. 185–194, 2020.

[6] A. H. Alshehri, “An online fake review detection approach using famous machine learning algorithms.” *Computers, Materials & Continua*, vol. 78, no. 2, 2024.

[7] M. Fayaz, A. Khan, J. U. Rahman, A. Alharbi, M. I. Uddin, and B. Alouffi, “Ensemble machine learning model for classification of spam product reviews,” *Complexity*, vol. 2020, no. 1, p. 8857570, 2020.

[8] G. Shahariar, S. Biswas, F. Omar, F. M. Shah, and S. B. Hassan, “Spam review detection using deep learning,” in *2019 IEEE 10th Annual Information Technology, Electronics and Mobile Communication Conference (IEMCON)*. IEEE, 2019, pp. 0027–0033.

[9] S. Mani, S. Kumari, A. Jain, and P. Kumar, “Spam review detection using ensemble machine learning,” in *Machine Learning and Data Mining in Pattern Recognition: 14th International Conference, MLDM 2018, New York, NY, USA, July 15-19, 2018, Proceedings, Part II 14*. Springer, 2018, pp. 198–209.

[10] K. Dhingra and S. K. Yadav, “Spam analysis of big reviews dataset using fuzzy ranking evaluation algorithm and hadoop,” *International journal of machine learning and cybernetics*, vol. 10, pp. 2143–2162, 2019.

[11] M. Lee, Y. H. Song, L. Li, K. Y. Lee, and S.-B. Yang, “Detecting fake reviews with supervised machine learning algorithms,” *The Service Industries Journal*, vol. 42, no. 13-14, pp. 1101–1121, 2022.

[12] W. Choi, K. Nam, M. Park, S. Yang, S. Hwang, and H. Oh, “Fake review identification and utility evaluation model using machine learning,” *Frontiers in artificial intelligence*, vol. 5, p. 1064371, 2023.

[13] N. Hussain, H. T. Mirza, I. Hussain, F. Iqbal, and I. Memon, “Spam review detection using the linguistic and spammer behavioral methods,” *IEEE Access*, vol. 8, pp. 53 801–53 816, 2020.

[14] S. Saumya and J. P. Singh, “Detection of spam reviews: a sentiment analysis approach,” *Csi Transactions on ICT*, vol. 6, no. 2, pp. 137–148, 2018.

[15] R. Barbado, O. Araque, and C. A. Iglesias, “A framework for fake review detection in online consumer electronics retailers,” *Information Processing & Management*, vol. 56, no. 4, pp. 1234–1244, 2019.